\pdfoutput=1

\documentclass[11pt]{article}

\usepackage[preprint]{acl}

\usepackage{times}
\usepackage{latexsym}

\usepackage[T1]{fontenc}

\usepackage[utf8]{inputenc}

\usepackage{microtype}

\usepackage{inconsolata}

\usepackage{graphicx}

\usepackage{amsfonts}
\usepackage{caption}
\usepackage{subcaption}
\usepackage{float}
\usepackage{booktabs}
\usepackage{multirow}
\usepackage{tabularx}
\newcommand{\eilev}{$\textbf{EILeV}$}

\title{Eliciting In-Context Learning in Vision-Language Models for Videos Through Curated Data Distributional Properties}

\author{Keunwoo Peter Yu \quad Zheyuan Zhang \quad Fengyuan Hu \quad Shane Storks \quad Joyce Chai \\
        Computer Science and Engineering Division \\
        University of Michigan \\
        Ann Arbor, MI, USA \\
        {\tt\{kpyu,zheyuan,hufy,sstorks,chaijy\}@umich.edu}}

\begin{document}
\maketitle
\begin{abstract}
A major reason behind the recent success of large language models (LLMs) is their \textit{in-context learning} capability, which makes it possible to rapidly adapt them to downstream text-based tasks by prompting them with a small number of relevant demonstrations. While large vision-language models (VLMs) have recently been developed for tasks requiring both text and images, they largely lack in-context learning over visual information, especially in understanding and generating text about videos. In this work, we implement \textbf{E}mergent \textbf{I}n-context \textbf{Le}arning on \textbf{V}ideos (\eilev{}), a novel training paradigm that induces in-context learning over video and text by capturing key properties of pre-training data found by prior work to be essential for in-context learning in transformers. In our experiments, we show that \eilev-trained models outperform other off-the-shelf VLMs in few-shot video narration for novel, rare actions. Furthermore, we demonstrate that these key properties of bursty distributions, skewed marginal distributions, and dynamic meaning each contribute to varying degrees to VLMs' in-context learning capability in narrating procedural videos. Our results, analysis, and \eilev{}-trained models yield numerous insights about the emergence of in-context learning over video and text, creating a foundation for future work to optimize and scale VLMs for open-domain video understanding and reasoning.\footnote{Code: \url{https://github.com/yukw777/EILEV}}
\end{abstract}
\section{Introduction}

In recent years, the advent of transformer-based~\cite{vaswani2017attention} large language models (LLMs) has garnered significant attention in and beyond the AI research community. A central reason for this is their \textit{in-context learning} capability \cite{brown2020language}, which makes it possible to rapidly adapt LLMs to novel tasks by simply prompting them with a few demonstrations. This capability removes the need for the expensive and arduous task-specific fine-tuning required by earlier language modeling approaches.

While in-context learning has been extensively studied and utilized in purely text-based problems in language understanding, reasoning, and generation, there are myriad potential applications for this rapid post-deployment adaptation in processing \textit{video}.
For example, in embodied and task-oriented AI, a major challenge is to recognize novel, rare human actions from video that cannot possibly be completely covered in training data \cite{perrett2023,successVQA,bao2023can}. A vision-language model (VLM) capable of in-context learning over video could address this challenge, as it would only require a few related videos of actions as few-shot, in-context examples to recognize and reason about these novel, rare actions.
However, while large VLMs for jointly processing text and images have been developed \cite{li2022blip,li2023blip-2,dai2023instructblip,zhu2023minigpt,peng2023kosmos,liu2023llava}, they are typically not optimized for reasoning over multiple images (i.e., frames), crucial for understanding videos. Meanwhile, a handful of open-source VLMs have recently been developed for video understanding \cite{zellers2022merlotreserve,li2023otter,zhang2023video,lin2023video}, but they lack in-context learning.

In-context learning in text-only, transformer-based LLMs was initially observed to improve with increased model size, along with the size and diversity of training data~\citep{brown2020language}. Later, \citet{chan2022data} identified several distributional properties of the training data as causes for this emergent behavior in transformer-based models: (1) bursty distributions with entities that tend to appear in clusters, (2) skewed marginal distributions with a long tail of infrequent items, and (3) dynamic meaning with label multiplicity. However, as their experiments relied on small transformer-based models trained on synthetic image classification data, it remains unclear whether their findings hold true for VLMs trained on video and text at scale. 

In this work, we address this question by conducting systematic empirical experiments to investigate whether these training data distributional properties also elicit in-context learning capabilities in VLMs for video.
Specifically, we use various text annotations from Ego4D~\citep{Ego4D2022CVPR}, a popular video dataset, to implement \textbf{E}mergent \textbf{I}n-context \textbf{Le}arning on \textbf{V}ideos (\eilev{}), a novel VLM training method that satisfies all three properties and successfully elicits in-context learning over video and text. In our experiments, we observe that the \eilev-trained models outperform other off-the-shelf VLMs in few-shot video narration on rare and out-of-distribution actions, and that, through careful ablation studies, each property indeed contributes to this in-context learning capability. Furthermore, our analysis yields a host of new insights around the importance of each property in in-context learning for video.

The contributions of our work are as follows: (1) we propose \eilev{}, a novel training method that can elicit in-context learning capabilities in VLMs for video and text, (2) we validate through systematic ablation experiments that the same data distributional properties that elicit in-context learning in small transformer-based models also apply to VLMs for videos, and (3) we release a set of \eilev{}-trained VLMs with in-context learning capabilities optimized for egocentric videos.

\section{Related Work}
\label{sec:related}

\subsection{In-Context Learning}

\citet{brown2020language} discovered in-context learning in LLMs when creating GPT-3. This was a significant departure from fine-tuning which involves parameter updates to adapt LLMs to downstream tasks. Instead, in-context learning enables LLMs to be adapted without parameter updates by prompting them with a few examples of a task as part of the input context for text generation. The size of the model and training data were thought to be key to training a model with in-context learning capabilities.

More recently, there has been more research on the exact causes of in-context learning. \citet{min2022metaicl} proposed MetaICL, a meta-training framework to elicit in-context learning capabilities in text-only language models. MetaICL conditions each example with related in-context examples during training. \citet{chan2022data} investigated the distributional properties of training data for in-context learning. Their findings showed that there are certain properties that encourage in-context learning in transformer-based models, and massive textual data from the web used to train LLMs naturally have those properties. Furthermore, \citet{reddy2023mechanistic} found that in-context learning is driven by the abrupt emergence of an induction head. There have also been works with findings about in-context learning in VLMs. Notably, training large generative VLMs with image-text interleaved data has been shown to be an effective technique to improve model performance, especially in tasks involving in-context learning \cite{alayrac2022flamingo,McKinzie2024MM1MA,Wang2024COSMOCS,tsimpoukelli2021multimodal,monajatipoor2023metavl}.
Our work combines these insights from prior work around the cause of in-context learning to propose a new VLM training paradigm for video and text, and carefully investigates how they contribute to in-context learning.

\subsection{Vision-Language Models (VLMs)}

With the recent success of text-only LLMs, there have been various efforts to replicate their success in multimodal settings, especially vision and language. Two different types of approaches in training generative VLMs have been proposed. The first is to train them from scratch using large text and paired image and text datasets \cite{hao2022language,huang2024language,peng2023kosmos,lu2023unified}. This approach allows the most controllability and flexibility as the resulting VLM is not dependent on other pre-trained models that may have undesirable behaviors, but it requires a massive amount of compute and data. In order to address these challenges, a number of approaches have been proposed to create VLMs by learning a mapping from a frozen pre-trained vision encoder to the input space of a frozen pre-trained LLM \cite{alayrac2022flamingo,li2023otter,zhao2023lavila,li2022blip,li2023blip-2,dai2023instructblip,liu2023llava,zhang2023video,lin2023video,yang2022frozenbilm,li2023videochat,zhu2023minigpt,laurenccon2023obelics,maaz2023video,ye2023mplug,gong2023multimodal,zhang2024llamaadapter}. 

Some of these approaches enable the resulting VLMs to process videos by representing them as sequences of still frames; however, only Flamingo~\citep{alayrac2022flamingo}, Otter~\citep{li2023otter} and Kosmos-2~\citep{peng2023kosmos} support in-context learning over video and text as a by-product of their large-scale pre-training. In this work, we conduct thorough investigation of how key properties of training data achieve in-context learning beyond just as a by-product of large-scale training.

\section{Three Distributional Properties for In-Context Learning}

Since \citet{brown2020language} discovered in-context learning in text-only LLMs, there has been much research into the cause for in-context learning. In particular, \citet{chan2022data} found that three characteristics of the training data are important in eliciting in-context learning in transformer-based models, each of which is abundant in both natural language and video data: \textit{bursty distributions}, \textit{skewed marginal distributions}, and \textit{dynamic meaning}.

\paragraph{Bursty Distributions}
In-context learning relies on data where entities appear in clusters, or non-uniformly depending on the context. Groups of related entities may be mentioned frequently in some contexts, but much more rarely in other contexts. This property is related to methods based on retrieval-augmented generation~\cite{lewis2020retrieval}.

\paragraph{Skewed Marginal Distributions}
In-context learning also relies on data of skewed marginal distributions with a long tail of infrequent items (i.e., a Zipfian distribution). 
This phenomenon is a long-standing challenge in representing language and images, and has long been observed in text, image, and video datasets collected for research.

\paragraph{Dynamic Meaning}
Lastly, in-context learning relies on dynamic meaning, where a single entity can have multiple possible interpretations, and multiple entities can map to the same interpretation. In natural language, we observe this property in word senses, homonyms, and synonyms. In the visual world, a particular object may be described in multiple valid ways, e.g., synonyms, physical properties, and hypernyms. Meanwhile, many distinct objects may be grouped based on various descriptors.

\section{Problem \& Methods}\label{sec:experimental design}

In this section, we first introduce the target problem and dataset for our evaluations of in-context learning. Next, we introduce \eilev{}, our training paradigm which captures all three distributional properties thought to elicit in-context learning, as well as the ablations we use to validate the importance of each property in enabling in-context learning over video and text. We then introduce the model architecture we apply this paradigm to, and lastly discuss how we evaluate the in-context learning capability of VLMs trained on video and text.

\subsection{Problem Definition}
We target the task of \textit{few-shot video narration} using the Ego4D dataset \cite{Ego4D2022CVPR}.

\paragraph{Few-Shot Video Narration}
\textit{Video narration} is a captioning task where given a video, a system must generate a text description of the events occurring in the video. Here, \textit{few-shot video narration} refers to the implementation of this task where a VLM (pre-trained on large-scale video and text data) is conditioned with one or more example videos and narrations before being prompted to generate a narration for a held-out video clip. If conditioning such a VLM on several example videos and narrations improves the quality of narration, this implies that the VLM is indeed capable of in-context learning over video and text.

\paragraph{Ego4D}
Ego4D is a popular large-scale dataset of egocentric videos that have been densely annotated with human-written English narrations, ideal for our task. Beyond narrations, the dataset includes higher-level class labels for the verbs and nouns associated with each narrated video clip. These annotations enable systematic ablations for all three distributional properties of training data discovered by \citet{chan2022data} to facilitate in-context learning, enabling a systematic study of in-context learning over video and text in VLMs. These ablations are introduced in Section~\ref{sec:ablations}.

\subsection{Training Paradigm \& Ablations}\label{sec:ablations}
\begin{figure*}[t]
    \centering
    \includegraphics[width=\linewidth]{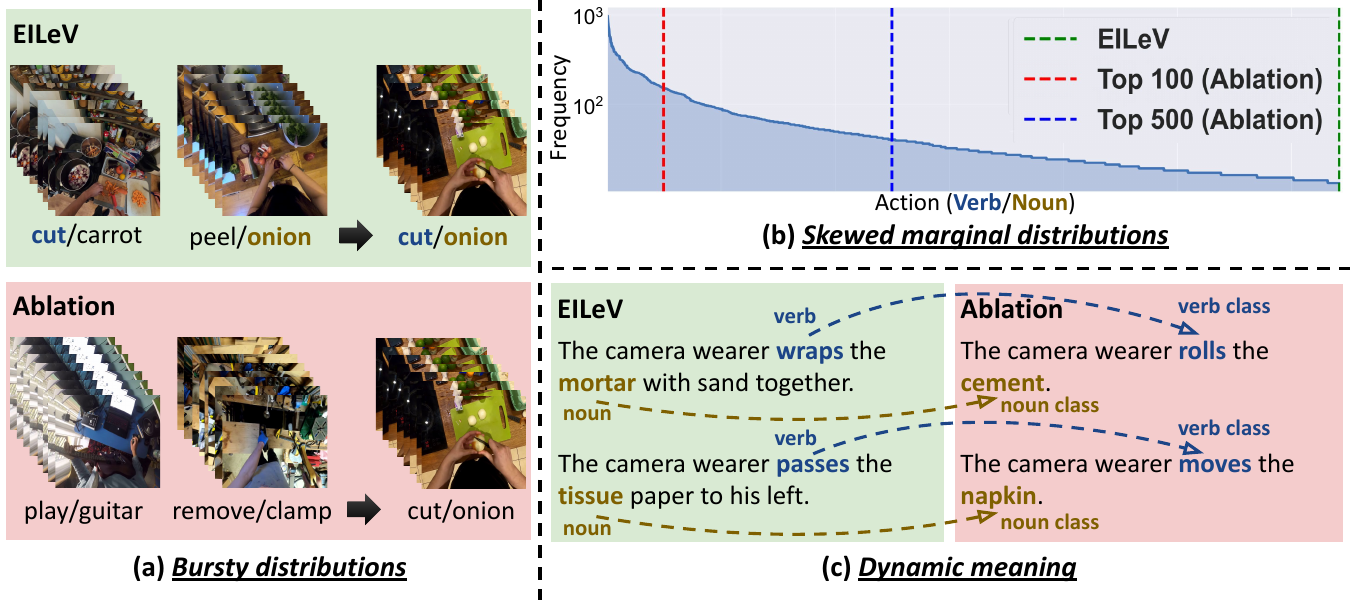}
    \caption{In our proposed training procedure \eilev{}, we ensure that the training data satisfy the following three properties: (a) bursty distributions, (b) skewed marginal distributions, and (c) dynamic meanings. Then, we ablate each property to demonstrate its importance. We ablate property (a) by randomly sampling in-context examples; we ablate property (b) by varying the number of common actions in the training data; we ablate property (c) by canonicalizing verbs and nouns using their corresponding verb and noun classes.}
    \label{fig:ablation}
\end{figure*}

Using Ego4D's ``Forecasting Hands \& Objects Master File'', we construct a dataset of interleaved text and video that satisfies these properties, and use it to train and evaluate VLMs. We call this training procedure \textbf{E}mergent \textbf{I}n-context \textbf{Le}arning on \textbf{V}ideos (\eilev{}).
\eilev{} uses the video and text data provided by Ego4D to implement all three distributional properties necessary for in-context learning: bursty distributions, skewed marginal distributions, and dynamic meaning. 
To demonstrate the importance of each distributional property captured in \eilev{}, we use Ego4D's detailed annotations to carefully ablate each property during training as illustrated in Figure \ref{fig:ablation}. Note that we ablate these properties only from the training data, not the evaluation data, and all of our models and baselines are given the same evaluation data with all of the distributional properties.

For all experiments, each training data point consists of a \textit{context} with 16 video-narration pairs, and a \textit{query} with a single video-narration pair. We convert the action narrations into question-answer pairs where the narrations are the answers, e.g., e.g., \textit{What is the camera wearer doing? The camera wearer cuts a carrot}.
We vary the syntactic form of questions using a set of templates (Appendix~\ref{sec:qa-template}). The training objective is to maximize the likelihood of the sequence of tokens in the ground-truth action narration, conditioned on the context and video clip from the query.

Next, we discuss how each distributional property was incorporated and ablated in \eilev{}.

\paragraph{Bursty Distributions}
In order to implement bursty distributions in \eilev{}, we take advantage of the annotations in Ego4D, where each video clip is annotated with a verb class and a noun class based on the main action portrayed in the clip. Specifically, we sample video clips and action narrations that share the same verb class as the query for half of the context, and we sample those with the same noun class for the other half. We further ensure that none of the sampled video clips and action narrations match both the verb class and noun class of the query simultaneously. This ensures that the context, while comprising a ``burst'' of similar concepts, only provides partial information regarding the query. This property can then be ablated by randomly sampling video clips and action narrations without regard to their verb and noun classes. Figure \ref{fig:ablation} (a) illustrates the two sampling strategies. We can measure the impact of bursty distributions by training VLMs with each type of context and comparing their in-context learning capabilities.

\paragraph{Skewed Marginal Distributions}
Like most natural datasets, Ego4D's verb and noun class labels have a skewed marginal distribution with a long tail of verb-noun pairs, making it ideal for our study. To study how the skewed marginal distributions of training data affect the in-context learning capability of trained models, we first use the verb and noun class annotations from Ego4D to designate the most frequent 80\% verb-noun pairs as \textit{common actions} for training, and the remaining 20\% as \textit{rare actions} only for evaluation. It is important to note that while none of the rare actions are part of the common action training data, they may still share either verb or noun classes with common actions. For example, if the training data contain common actions (\textit{put}, \textit{key}) and (\textit{sit}, \textit{bench}), there may exist a rare action (\textit{put}, \textit{bench}) in the evaluation data. 

To measure how the skewness of marginal distributions in the training data impacts models' capability to generalize to these novel held-out actions, we then vary the number of common actions in the training data through three experiments. Specifically, we construct a training dataset with only the top 100 common actions (little skewness without a long tail of infrequent actions), one with the top 500 common actions (moderate skewness with a short tail of infrequent actions) and another with all the common actions (highly skewed with a long tail of infrequent items). We uniformly upsample the datasets with top 100 and top 500 common actions to keep all three training datasets to be the same size. Figure \ref{fig:ablation} (b) shows how these training datasets with different marginal distributions are constructed. Given these curated training datasets, we can measure the impact of the skewness of the marginal distributions of the training data on trained models' in-context learning capability.

\paragraph{Dynamic Meaning}
For dynamic meaning, we rely on the fact that Ego4D's natural language action narrations contain words of multiple senses, homonyms, and synonyms. To ablate this dynamic meaning property in \eilev{}, we canonicalize verbs and their corresponding objects in the action narrations. Specifically, we prompt an LLM (Llama-2-Chat 7B; \citealp{touvron2023llama2}) to replace the verb and its corresponding object of each action narration with their verb and noun class. Figure \ref{fig:ablation} (c) shows the canonicalization process. We can then measure the impact of dynamic meaning by comparing the in-context learning capability of VLMs trained on data with and without this property.

\subsection{Model}
\label{subsec:model}
To experiment with \eilev{} as discussed above, we adopt a VLM architecture capable of processing sequential data interleaved with both video clips and texts, making it possible to infer patterns and relationships among them and thus support the emergence of in-context learning over them. We initialize our model with BLIP-2~\citep{li2023blip-2}, a VLM created by learning a transformer-based projection (called a querying transformer or Q-Former) from a frozen pre-trained vision encoder into the input space of a frozen LLM. Since BLIP-2's original implementation is not able to handle data interleaved with video clips and texts, we follow \citet{hao2022language} to perform simple modifications to enable its frozen language model to serve as a universal interface for video clips and texts.\footnote{While there exist VLMs that already natively support interleaved video and text \citep{alayrac2022flamingo,awadalla2023openflamingo,li2023otter}, we intentionally chose a VLM that did not to isolate the impact of our \eilev{} training paradigm on VLMs' in-context learning capability.} 
Specifically, we first encode all the video clips by independently encoding sampled frames with BLIP-2's frozen Vision Transformer (ViT)-based~\cite{dosovitskiy2021an} vision encoder to produce a sequence of vision tokens for each video clip. The sequence of vision tokens is then compressed by BLIP-2's Q-Former into a fixed-length sequence. The fixed-length sequence is further projected to the word embedding space of the frozen language model of BLIP-2 by a linear layer. It is then interleaved with the text tokens according to the order in which video clips and texts appear in the interleaved data to form the input to the frozen language model. Following the fine-tuning procedure of \citet{li2023blip-2}, we freeze the vision encoder and language model of the BLIP-2 models during training. For all of our experiments, we use BLIP-2 with 2.7 billion parameter OPT~\cite{zhang2022opt} as its frozen language model (BLIP-2 OPT-2.7B), and BLIP-2 with XL-size Flan-T5~\cite{wei2022finetuned} as its frozen language model (BLIP-2 Flan-T5-xl).\footnote{We intentionally use the smaller BLIP-2 variants in order to remove the model size as a confounding variable for in-context learning.}

\subsection{Evaluation}\label{sec:evaluation}
To evaluate our various model ablations, we need a means to measure the quality of action narrations generated by models, and the degree to which in-context learning supports this generation.

\subsubsection{Action Narration Generation}
One major difficulty in evaluating generative models for the action narration generation task is that there is no single correct way to describe the action in a video clip.
In an ideal world, we would rely on human annotators to rate how close a generated action narration is to the ground truth, but the cost to do so would be prohibitive. In order to address this challenge, a number of semantic-similarity-based metrics~\cite{zhang2019bertscore,reimers-2019-sentence-bert} that correlate closely with human judgment have been proposed, and we take advantage of them in our evaluations. Specifically, we report the performance along semantic similarity-based scores produced by Siamese Sentence-BERT Bi-Encoder (STS-BE; \citealp{reimers-2019-sentence-bert}).
For completeness, we also report ROUGE-L~\cite{lin2004rouge}, a lexical-based text generation metric.

\subsubsection{In-Context Learning Capability}
To evaluate the in-context learning capability of trained models for action narration, we vary the number of in-context examples in context-query instances (different numbers of ``shots'') and calculate the above text generation metrics for generated action narrations on the test set. If adding more shots improves narration quality under these metrics, this suggests that the VLM is successfully using in-context learning to adapt to the action narration generation task. Within a single experiment setting, we use the same pre-sampled in-context examples with all of the three distributional properties to ensure fair comparison.

\section{Experimental Results}

In our experiments, we find that the performance of both \eilev-trained models strictly increases as more in-context examples (shots) are provided, indicating that \textbf{our models successfully acquired in-context learning capabilities during training.}
First, in Section~\ref{subsec:off-the-shelf-vlms}, we establish the in-context learning capability of our models by measuring their performance on rare actions they were not trained on (the key challenge motivating this work), and compare their performance to that of off-the-shelf VLMs. In Section~\ref{subsec:out-of-dist-actions}, we confirm our models' ability to generalize to out-of-distribution actions via in-context learning without fine-tuning by evaluating their performance on such actions. In Sections~\ref{subsec:results-bursty}, \ref{subsec:results-skewed}, and \ref{subsec:dynamic-meaning}, we compare their performance to that of models trained on datasets with each key distributional property ablated (as described in Section~\ref{sec:ablations}) to explore the impact of these training data properties on in-context learning for video and text in VLMs.

\subsection{Generalization to Rare Actions}
\label{subsec:off-the-shelf-vlms}

\begin{table}
\centering
\begin{tabularx}{\linewidth}{@{}ll@{}}
\toprule[2pt]
Model                                                       & MMI Dataset Size                                                                   \\ \midrule[2pt]
\multirow{3}{*}{\shortstack[l]{\eilev~BLIP-2\\OPT-2.7B \&\\ Flan-T5-xl}} & \multirow{2}{*}{\shortstack[l]{115K context-query\\instances}}        \\
                                                            &                                                                                    \\
                                                            &                                                                                    \\ \midrule
Kosmos-2                                                    & \multirow{2}{*}{\shortstack[l]{71M image-text\\webpages~\citep{huang2023language}}} \\ 
                                                            &                                                                                    \\ \midrule
                                                            & \multirow{4}{*}{\shortstack[l]{101.2M image-text\\webpages~\citep{zhu2023multimodal} \&\\2.8M context-query\\instances~\citep{li2023mimic}}}  \\ 
Otter                                                       &                                                                                    \\ 
                                                            &                                                                                    \\ 
                                                            &                                                                                    \\ \bottomrule[2pt]
\end{tabularx}
\caption{Off-the-shelf and \eilev-trained VLMs and their multi-modal interleaved (MMI) dataset sizes.}
\label{tab:models}
\end{table}

We first compare our \eilev-trained models with existing off-the-shelf VLMs in the challenging practical setting that motivated this work: \textit{adaptation to rare actions}. Specifically, we evaluate our models, Kosmos-2~\cite{peng2023kosmos}, and Otter~\cite{li2023otter} on the evaluation set of held-out rare action videos from Ego4D described in Section \ref{sec:ablations}.\footnote{Our models were not trained on these rare actions, and Kosmos-2 was not trained on Ego4D. While Otter was trained on Ego4D, the video-text training data was not interleaved as proposed for \eilev-trained models, and the low frequency of these actions nevertheless poses a significant challenge.}

We choose these two models as they are the only open-source large VLMs that support video input and in-context-learning out-of-the-box at the time of writing. Furthermore, we purposely exclude proprietary models like GPT-4~\cite{achiam2023gpt} or Gemini~\cite{team2023gemini} as we cannot verify if they truly perform in-context learning over videos and texts under the hood (they may use complex data preprocessing pipelines that involve many auxiliary steps like OCR). Compared to our \eilev-trained models, these models have been trained on far more multi-modal interleaved (MMI) data directly related to in-context learning over video (Table \ref{tab:models}), as well as other naturalistic multi-modal and text data from the Internet. They also have far more trainable parameters: Kosmos-2 has 1.6 billion and Otter has 1.3 billion, while our models have 188 million (the same number as BLIP-2). Further, unlike our architectural modification that represents each video with a fixed-length sequence, Kosmos-2 and Otter both treat each video as a sequence of images. 
For an evaluation representative of the practical usage of VLMs, we do not fine-tune models (which requires prohibitive computing power). Instead, we rely solely on models' in-context learning capability to adapt to these rare actions.

\begin{figure}
    \centering
    \includegraphics[width=\linewidth]{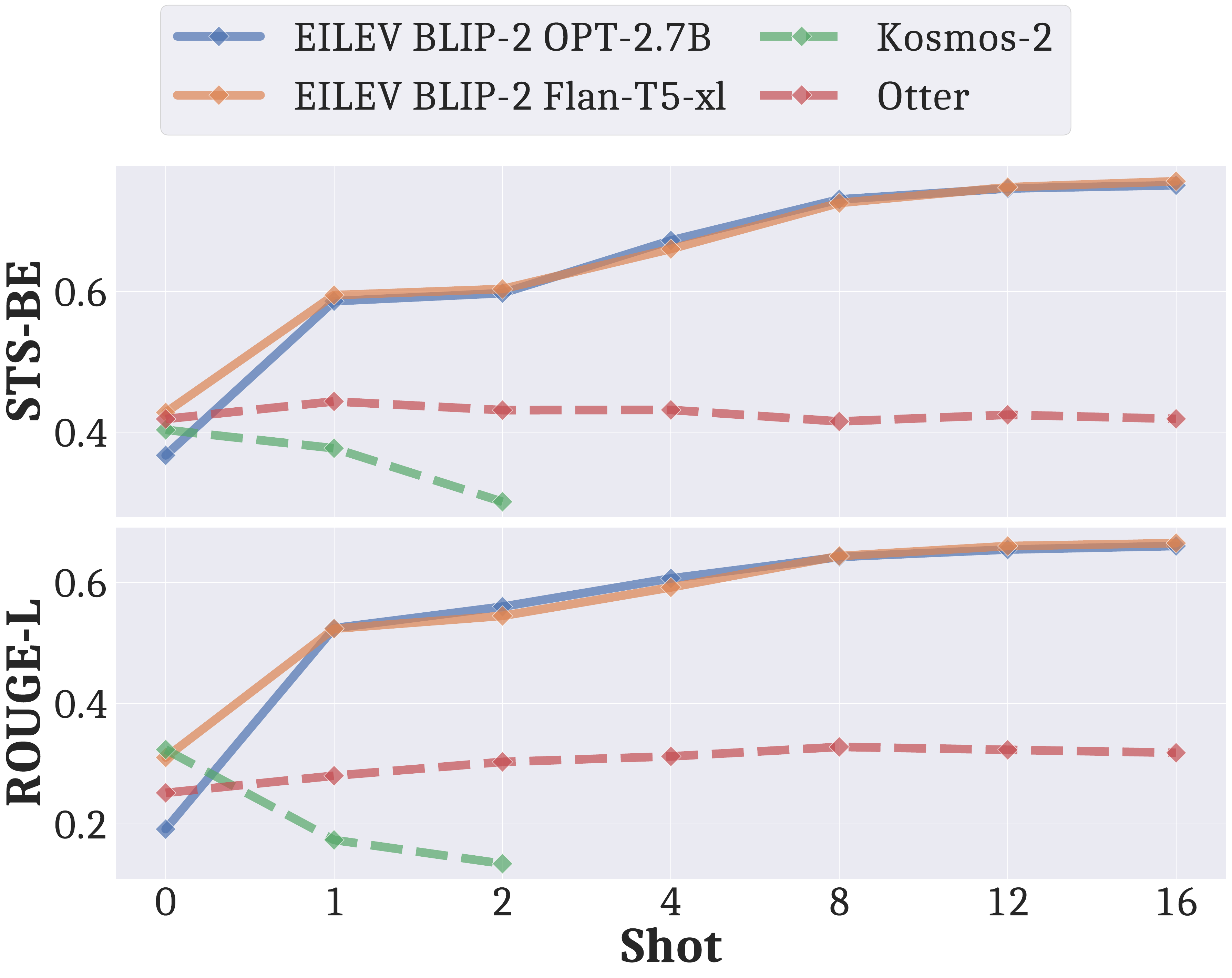}
    \caption{Performance of \eilev-trained and off-the-shelf VLMs (Kosmos-2 and Otter) on the evaluation set of held-out rare actions from Ego4D.}
    \label{fig:off-the-shelf-vlms}
\end{figure}

Figure \ref{fig:off-the-shelf-vlms} shows the results of this evaluation.\footnote{We can only perform evaluations up to 2-shot with Kosmos-2, as it runs out of its context window beyond 2-shot.} While the zero-shot performance of our \eilev-trained models is similar to Kosmos-2 and Otter, \textbf{as we provide in-context examples, the performance of our models increases while that of off-the-shelf VLMs does not}. Consequently, our \textbf{\eilev-trained VLMs significantly outperform off-the-shelf VLMs}. While Kosmos-2 and Otter have not been fine-tuned on this exact data, they are much larger models trained on an enormous amount of naturalistic data, and their in-context learning capability is a main selling point thought to remove the need for task-specific fine-tuning. 
Therefore, it is reasonable to expect their performance to improve with more in-context examples or even outperform our models. This observation underscores that \textit{training smaller VLMs with a focused approach like \eilev{} can be more advantageous for certain use-cases}, such as generating narrations for novel, rare actions, than training large, generalist VLMs on huge naturalistic datasets.

\subsection{Generalization to Out-of-Distribution Actions}
\label{subsec:out-of-dist-actions}

\begin{figure}[b]
    \centering
    \includegraphics[width=\linewidth]{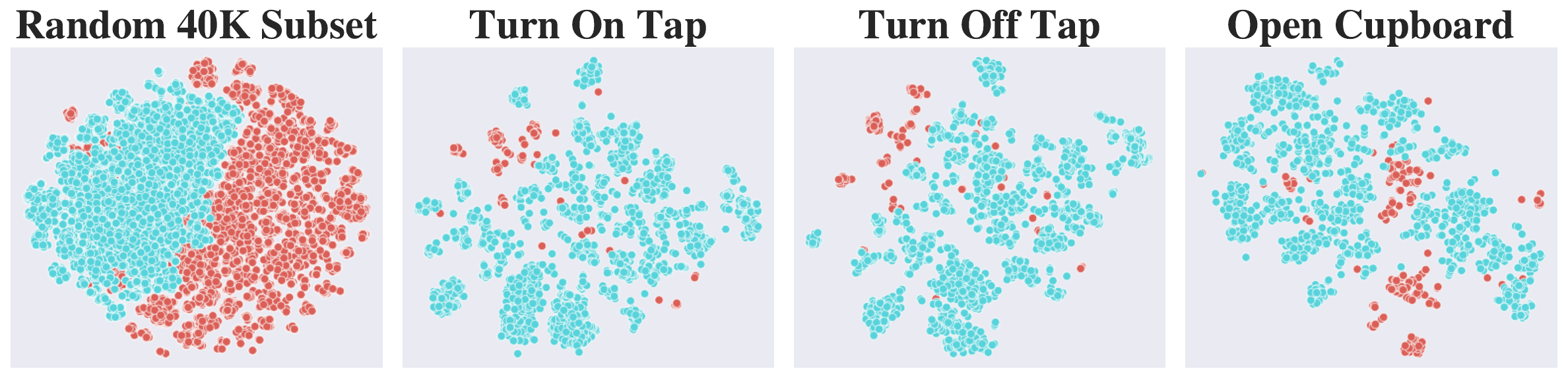}
    \caption{t-SNE plots of the video embeddings from the frozen vision encoder of BLIP-2 OPT-2.7B. Ego4D videos are in \textcolor[HTML]{db5f57}{red}, and EPIC-KITCHENS-100 videos are in \textcolor[HTML]{57d3db}{blue}. Plots for a randomly sampled subset of 40k videos from both and three most common actions from EPIC-KITCHENS-100 are shown. We manually map Ego4D actions to the EPIC-KITCHENS-100 actions.}
    \label{fig:tsne}
\end{figure}

Next, we test if \eilev-trained BLIP-2 models trained solely on Ego4D can generalize to out-of-distribution actions via in-context learning. Specifically, we evaluate them on the validation split of a different egocentric video dataset, EPIC-KITCHENS-100~\cite{Damen2022RESCALING}, without further fine-tuning. Note that there is a significant distributional shift between Ego4D and EPIC-KITCHENS-100 even though they both contain egocentric videos in the kitchen setting as evidenced by the t-SNE plot in Figure \ref{fig:tsne}. All the experimental setups are same as Section \ref{subsec:off-the-shelf-vlms} except the evaluation context-query instances are formed by sampling both the context and the query from the validation set of EPIC-KITCHENS-100 with all three distributional properties. Unlike Ego4D, the action narrations from EPIC-KITCHENS-100 are not full sentences, but simple verb-noun phrases. Therefore, we use an LLM (7 billion parameter Llama-2-Chat~\cite{touvron2023llama2}) to turn the simple verb-noun phrases into full sentences with ``the camera wearer'' as the subject.

\begin{figure}
    \centering
    \includegraphics[width=\linewidth]{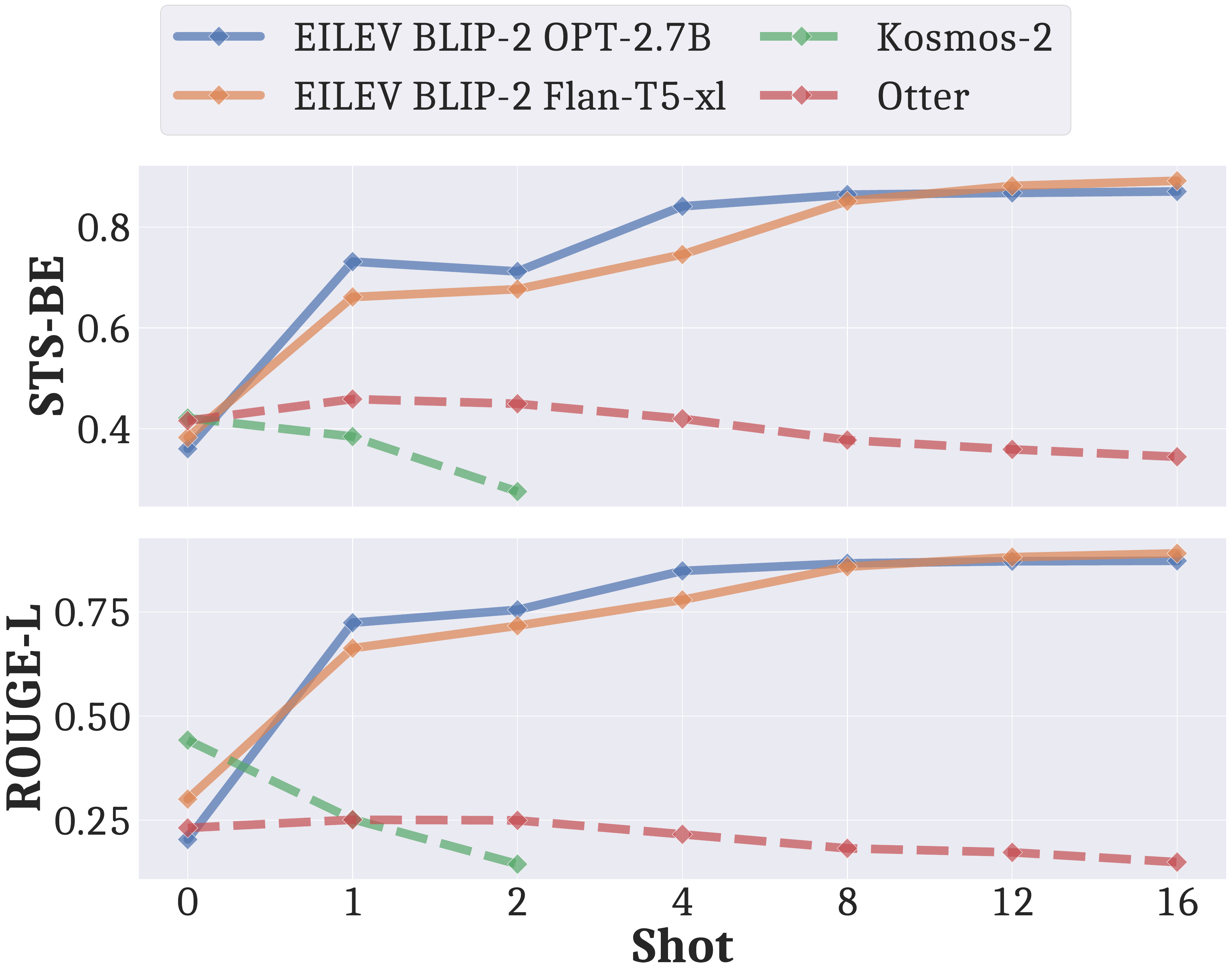}
    \caption{Performance of \eilev-trained and off-the-shelf VLMs (Kosmos-2 and Otter) on the validation set of out-of-distribution actions from EPIC-KITCHENS-100.}
    \vspace{-10pt}
    \label{fig:off-the-shelf-vlms-ek}
\end{figure}

Figure \ref{fig:off-the-shelf-vlms-ek} reports the evaluation results. \textbf{The performance of the \eilev-trained BLIP-2 models improves with an increasing number of in-context examples, ultimately outperforming all the baselines.} Similar to the trends observed on the Ego4D-based dataset, all baseline models demonstrate comparable performance in the 0-shot setting but fail to benefit from in-context examples, resulting in our \eilev-trained models outperforming them. These results further support that \textit{training smaller VLMs with a targeted approach like \eilev~can be more advantageous--even for generating narrations of out-of-distribution actions--than training large, generalist VLMs on extensive naturalistic datasets.}

\subsection{Bursty Distributions Ablation}
\label{subsec:results-bursty}

\begin{figure}
    \centering
    \includegraphics[width=\linewidth]{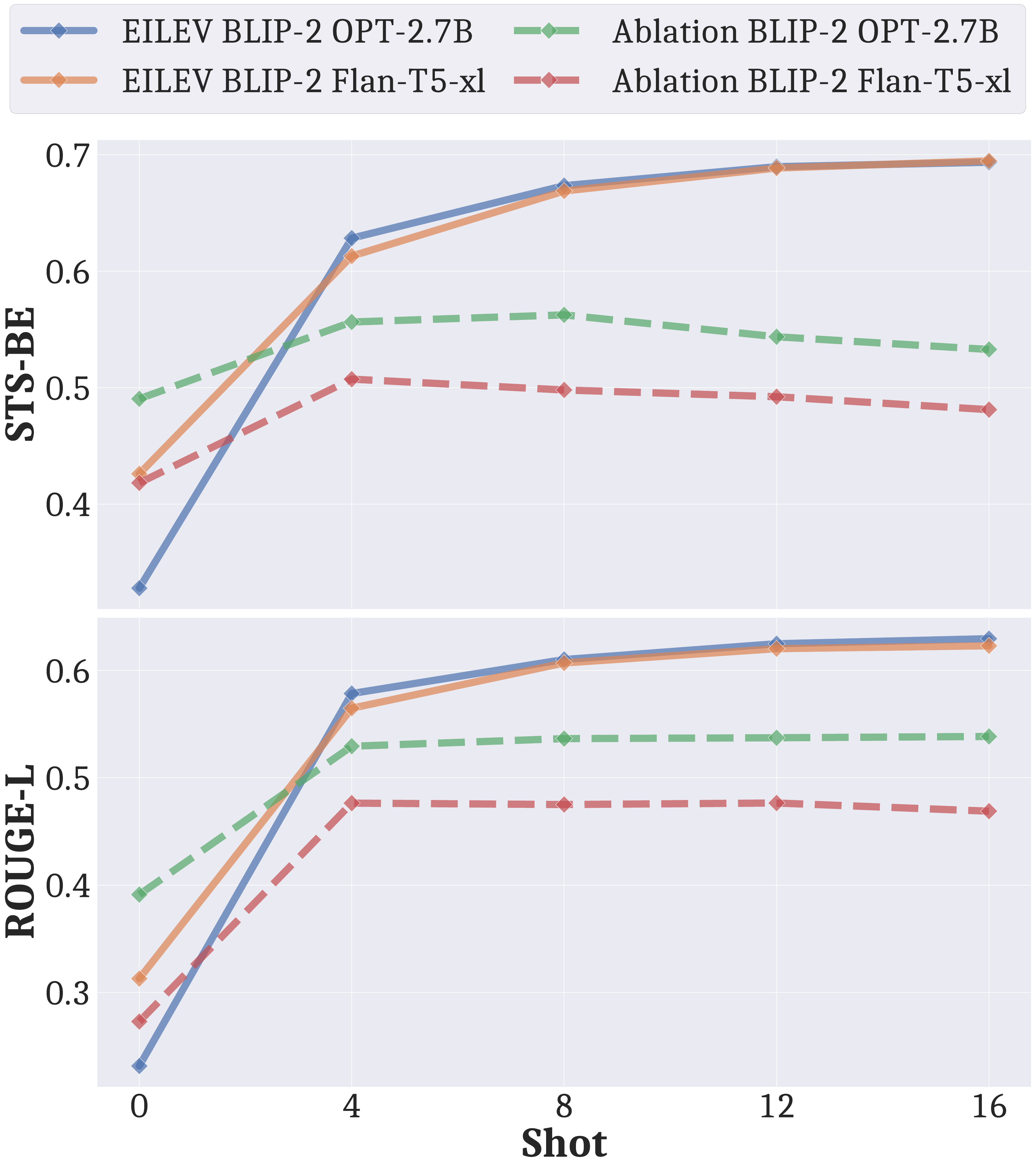}
    \caption{Results for the bursty distributions ablation experiment.}
    \label{fig:bursty-ablation}
\end{figure}

Figure \ref{fig:bursty-ablation} shows the results of the bursty distributions ablation experiment. To maintain the same action distributions in both the training and test sets, we use a random train-test split with a ratio of 75/25 for this experiment. Unlike the \eilev-trained models, the performance of the models trained on randomly sampled in-context examples (ablation) initially improves from 0-shot to 4-shot, but tapers or even decreases as more examples are provided. This indicates that they failed to acquire in-context learning capabilities during training, suggesting that \textbf{bursty distributions are indeed necessary for in-context learning on video and text}. We hypothesize that the initial improvement in performance from 0-shot to 4-shot is mainly due to the fact that ablation models have learned to mimic lexical characteristics from in-context examples. However, as they have failed to learn to exploit the semantic information from in-context examples due to the lack of bursty distributions in training data, they do not benefit from additional in-context examples.

\begin{figure}[ht]
    \centering
    \includegraphics[width=\linewidth]{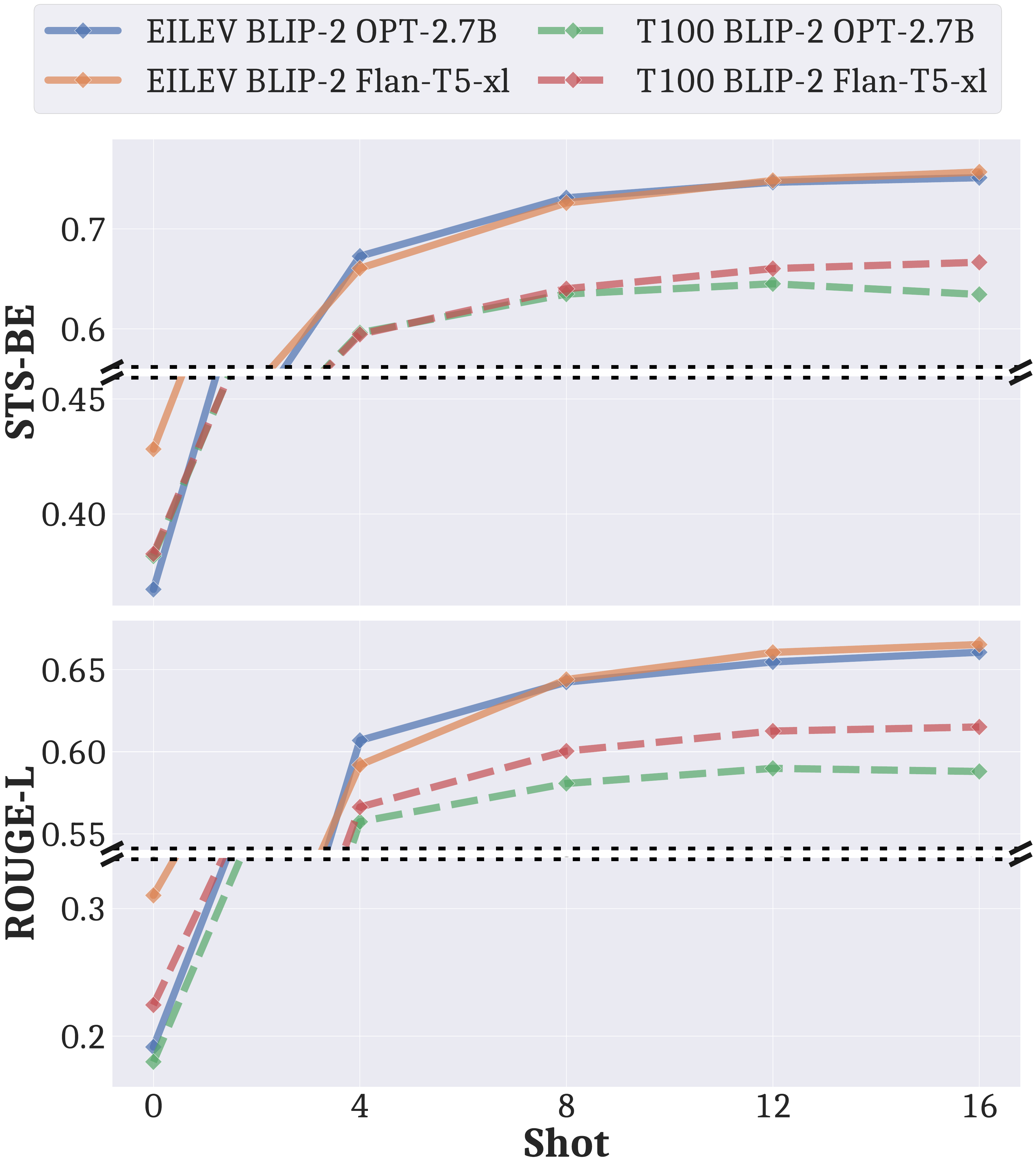}
    \caption{Results for the skewed marginal distributions ablation experiment using a training dataset with top 100 common actions (T100).}
    \label{fig:skewed-t100-ablation}
\end{figure}

\begin{figure}[ht]
    \centering
    \includegraphics[width=\linewidth]{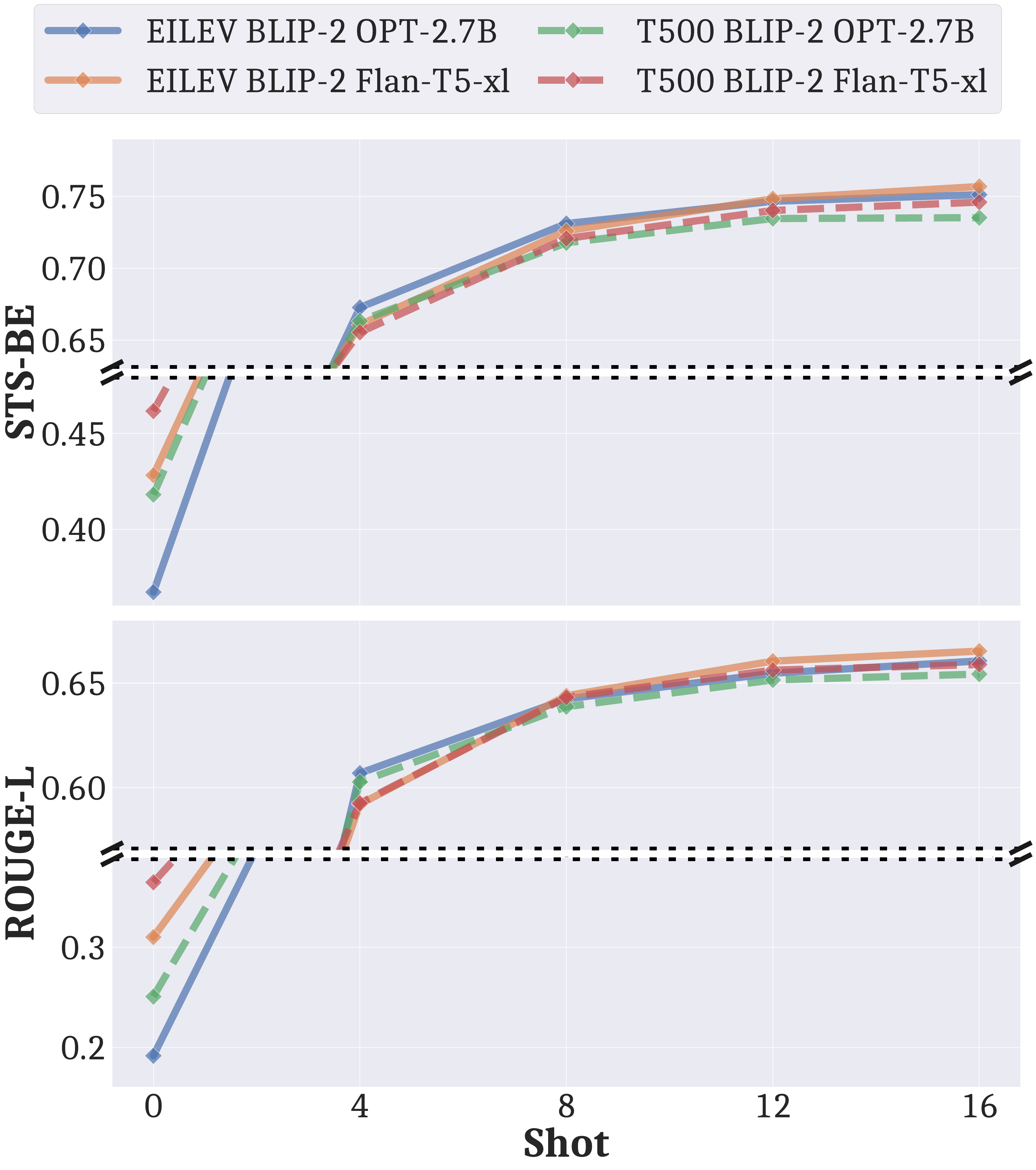}
    \caption{Results for the skewed marginal distributions ablation experiment using a training dataset with top 500 common actions (T500).}
    \label{fig:skewed-t500-ablation}
\end{figure}

\subsection{Skewed Marginal Distributions Ablation}
\label{subsec:results-skewed}

Figures \ref{fig:skewed-t100-ablation} and \ref{fig:skewed-t500-ablation} show the results of the skewed marginal distribution ablation experiment. The T100 models trained on data with only the top 100 common actions (little skewness without a long tail of infrequent actions) show a noticeably inferior in-context learning performance to the \eilev-trained models that were trained on the training dataset with all the common actions (highly skewed with a long tail of infrequent items). On the other hand, the T500 models trained on data with the top 500 common actions (moderate skewness with a short tail of infrequent actions) show an in-context learning performance that is only slightly worse than the \eilev-trained models, indicating that \textbf{an increased amount of skewness with a long tail of infrequent items makes in-context learning more likely to appear in VLMs}. Further, we observe that the T500 models outperform their respective \eilev-trained models in the 0-shot setting. This is an instance of in-context versus in-weights learning tradeoff (also studied in \citealp{chan2022data}), a phenomenon where in-context learning capability can reduce pre-trained models' ability to utilize knowledge encoded in their weights during pre-training. Interestingly, we do not observe this pattern with the T100 models, perhaps because the less diverse training data is not representative enough for models to gain sufficient in-weights knowledge.

\subsection{Dynamic Meaning Ablation}
\label{subsec:dynamic-meaning}

\begin{figure}
    \centering
    \includegraphics[width=\linewidth]{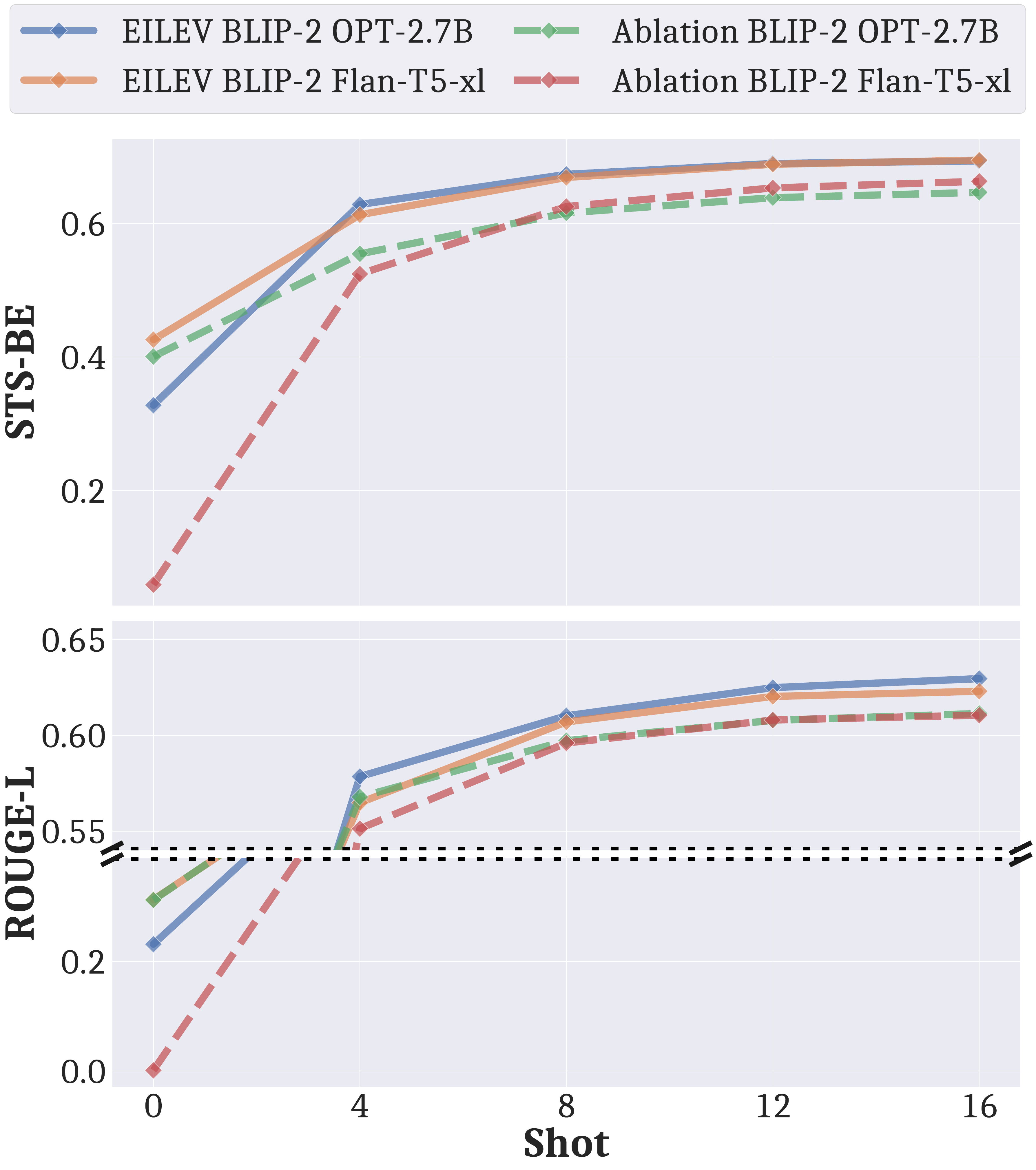}
    \caption{Results for the dynamic meaning ablation experiment.}
    \label{fig:dynamic-ablation}
\end{figure}

Figure \ref{fig:dynamic-ablation} shows the results of the dynamic meaning ablation experiment. We use a random train-test split with a ratio of 75/25 for this experiment to maintain the same action distributions in both the training and test sets. The ablation models trained on data with verbs and their corresponding objects canonicalized surprisingly acquire some in-context learning capabilities, but the \eilev-trained models mostly outperform them. Since the performance gaps under this ablation are smaller than that of the previous ablations, this suggests that \textbf{while dynamic meaning plays a role in the in-context capabilities of a VLM, it contributes less than bursty and skewed marginal distributions do}. Interestingly, however, the performance gap is much more pronounced for STS-BE (semantic similarity metric) than ROUGE-L (lexical metric), suggesting that dynamic meaning contributes more to the model's ability to extract semantic information from in-context examples than lexical information.

\section{Conclusion}

In this work, we conducted a first-of-its-kind systematic investigation of in-context learning in vision-language models (VLMs) trained on videos and text. Specifically, we implemented \textbf{E}mergent \textbf{I}n-context \textbf{Le}arning on \textbf{V}ideos (\eilev{}), a novel training paradigm capturing three key properties of training data found to induce in-context learning in transformers \cite{chan2022data}: bursty distributions, skewed marginal distributions, and dynamic meaning. In our experiments, we showed that our \eilev-trained models exhibit in-context learning capabilities superior to that of off-the-shelf VLMs, as they were significantly more adaptable to novel, rare actions, as well as out-of-distribution actions. We demonstrated that all three of these properties are indeed important to optimize the in-context learning capabilities of these models on narrating actions in videos, especially bursty and skewed marginal distributions. 

Our work yields new insights about the nature of in-context learning in video and text. For example, we observed that while reducing the skewness of the training data distribution compromised in-context learning capability, it improved in-weights learning in trained models \cite{chan2022data}. We also found that dynamic meaning had a bigger impact on semantic similarity metrics for generated narrations than lexical metrics, suggesting this property is particularly important for acquiring semantic information through in-context learning.

While we focused on action narration in Ego4D \citep{Ego4D2022CVPR} as a proof-of-concept, \eilev{} serves as a foundation for the community to build VLMs capable of in-context learning on video and text in broader tasks and domains. We release our \eilev{}-trained models as a resource for future work in egocentric video narration.

\section*{Limitations}

Since our \eilev-trained models are optimized and evaluated for action narration generation on egocentric video using in-context learning, their ability to generalize to diverse, real-world scenarios may be limited. However, this focus was by design and necessity. The primary goal of this work was to verify that the three distributional properties identified by \citet{chan2022data} also elicit in-context learning capabilities in VLMs for videos. To that end, we intentionally chose to use Ego4D, a dataset with sufficient annotations to enable our systematic ablation experiments as a proof of concept. Despite this limitation, \eilev-trained models may retain some capability to answer other types of questions due to the use of a frozen language model. Furthermore, \eilev{} is a general training method that can be applied to other tasks given the appropriate data.

Additionally, our models may inherit biases from their frozen language models, making it possible that they could generate harmful content. Before deploying such a system for real-world applications, safety measures like guardrails and training data sanitization are crucial to minimize potential negative impact. On the other hand, since we used the diverse and global data from Ego4D to train our models, this may mitigate possible socio-economic bias found in pre-trained visual representations \cite{nwatu-etal-2023-bridging}.

\section*{Acknowledgments}
This work was supported in part by the DARPA PTG program HR00112220003. We would like to thank the entire MSPR team for their helpful discussion and feedback. We would also like to thank the anonymous reviewers for their valuable comments and suggestions.

\bibliography{custom}

\appendix
\section{Additional Experiments}

\subsection{Additional Baselines}
\label{subsec:additional-baselines}

We report the performance of three additional baselines on the Ego4D-based dataset used in the main ablation experiments, as well as another dataset constructed from EPIC-KITCHENS-100. The first is a naive action classification baseline (``VideoMAE''). Specifically, we fine-tune the ``videomae-huge-finetuned-kinetics'' variant of VideoMAE~\citep{tong2022videomae} using the verb and noun class annotations to produce a verb and a noun classifier. The predicted verb and noun classes are then transformed into action narrations using an off-the-shelf LLM (7 billion parameter Llama-2-Chat~\citep{touvron2023llama2}). Note that this baseline only uses videos as its input, and cannot perform in-context learning. The second are off-the-shelf BLIP-2 models with the architectural modifications from Section \ref{subsec:model} for interleaved data support (``BLIP-2 OPT-2.7B \& Flan-T5-xl''). The third are \eilev-trained models with in-context examples ablated, and fine-tune solely on the query (``FT BLIP-2 OPT-2.7B \& Flan-T5-xl'').

\subsubsection{Results on Ego4D}

\begin{figure}[ht]
    \centering
    \begin{subfigure}{\linewidth}
        \centering
        \includegraphics[width=\linewidth]{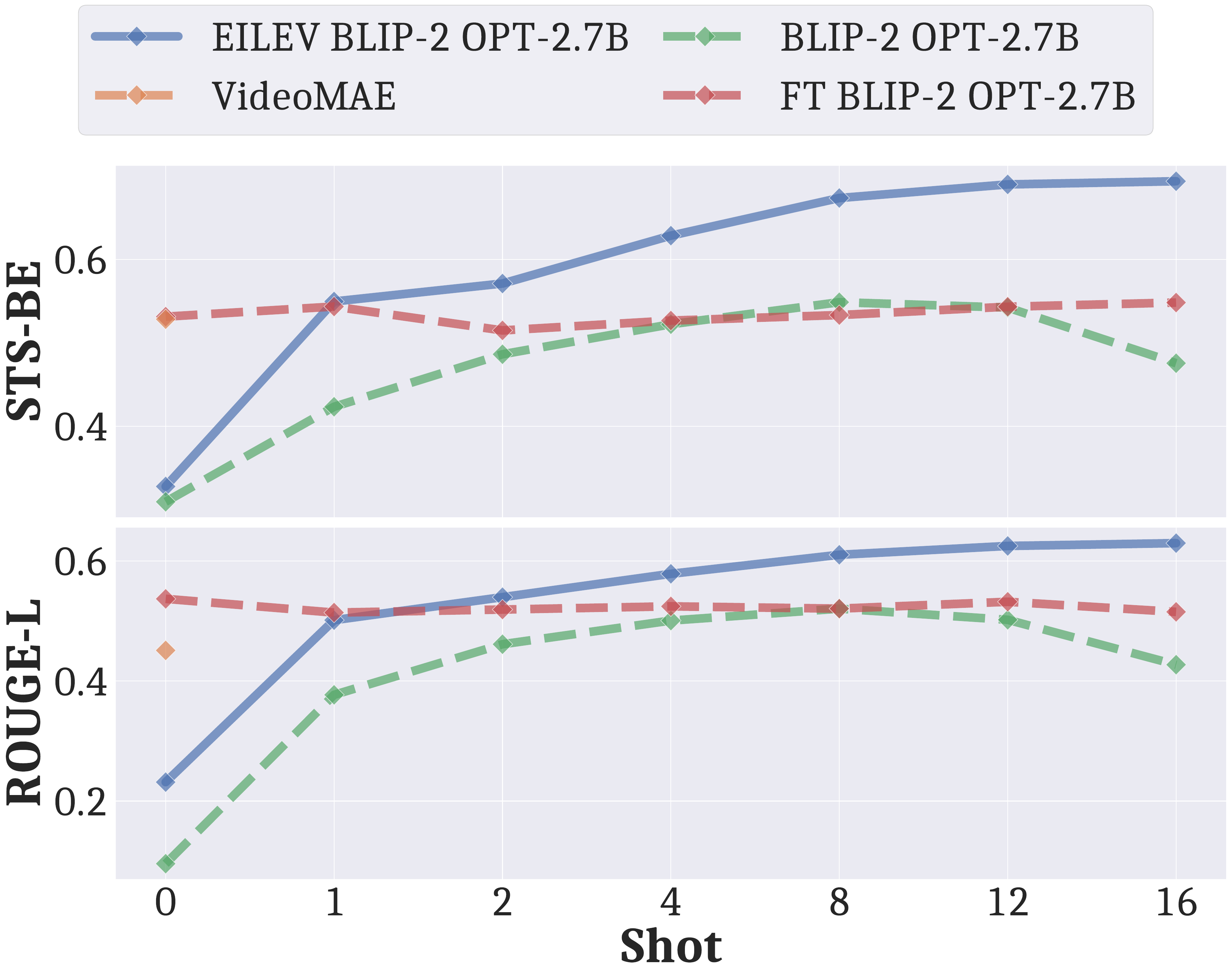}
    \end{subfigure}
    \begin{subfigure}{\linewidth}
        \centering
        \includegraphics[width=\linewidth]{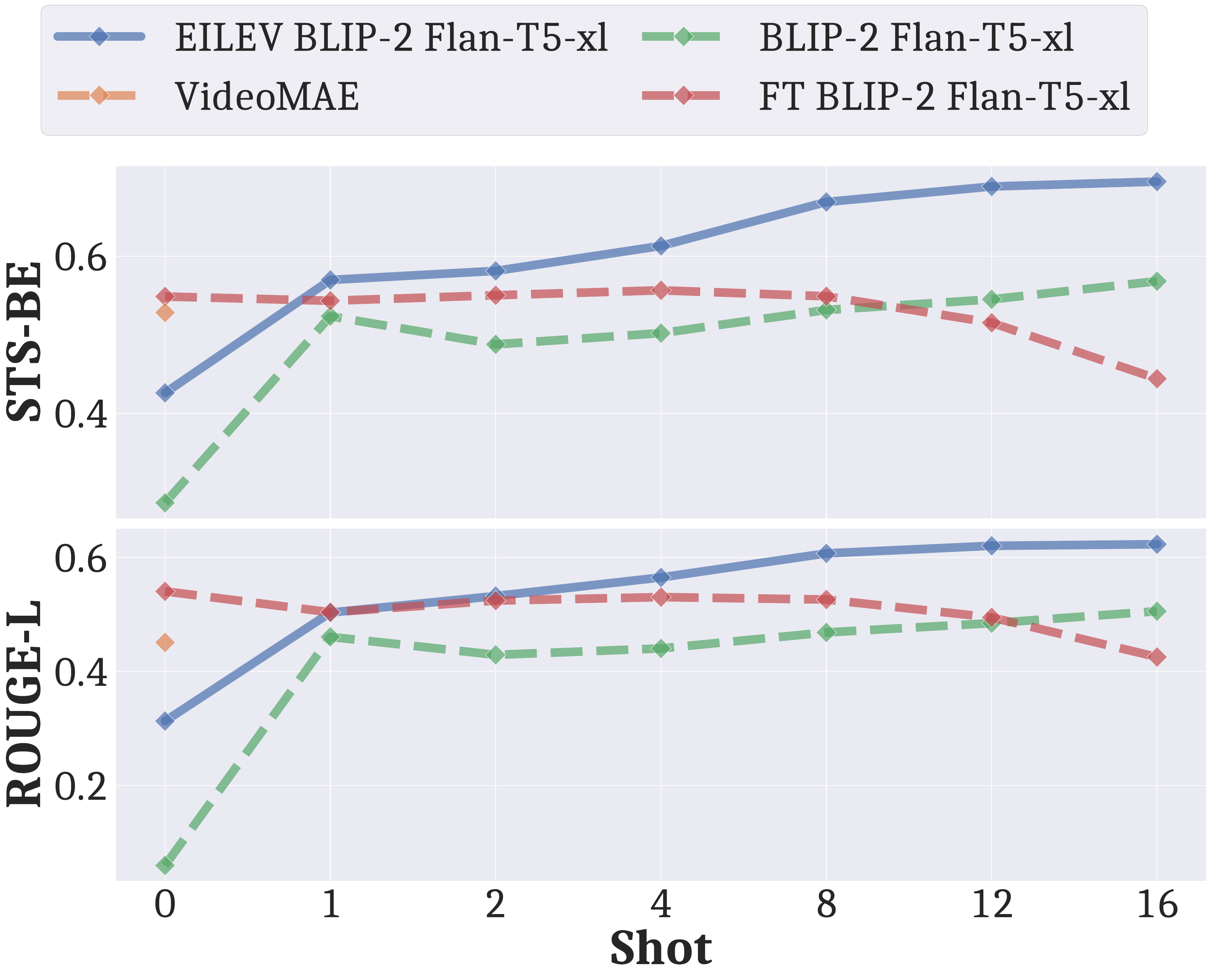}
    \end{subfigure}
    \caption{Performance of additional baselines on the Ego4D-based dataset.}
    \label{fig:ego4d-eval}
\end{figure}

Figure \ref{fig:ego4d-eval} reports the performance of the three additional baselines on the Ego4D-based dataset. The VideoMAE and FT BLIP-2 models exhibit the best performance at 0-shot, suggesting they have the most amount of in-weights knowledge due to their fine-tuning. However, VideoMAE cannot process in-context examples, and its 0-shot performance is quickly outperformed by \eilev-trained models with only one in-context example. The performance of FT BLIP-2 models stagnates or even declines as the number of shots increases, highlighting their lack of in-context learning capabilities and the importance of the training data design discussed in Section \ref{sec:ablations}. These findings about the performance of different models at 0-shot and subsequent shots align with \citet{chan2022data} observations regarding the ``tradeoff between in-context learning and in-weights learning,'' where no models could maintain both in their experiments. In our experiment, the \eilev-trained BLIP-2 models are optimized for in-context learning, as evidenced by their subpar performance at 0-shot and superior performance with additional shots, whereas the FT BLIP-2 models show the opposite trend. We leave designing training data to find the right balance for future work.

\subsubsection{Results on EPIC-KITCHENS-100}

\begin{figure}[ht]
    \centering
    \begin{subfigure}{\linewidth}
        \centering
        \includegraphics[width=\linewidth]{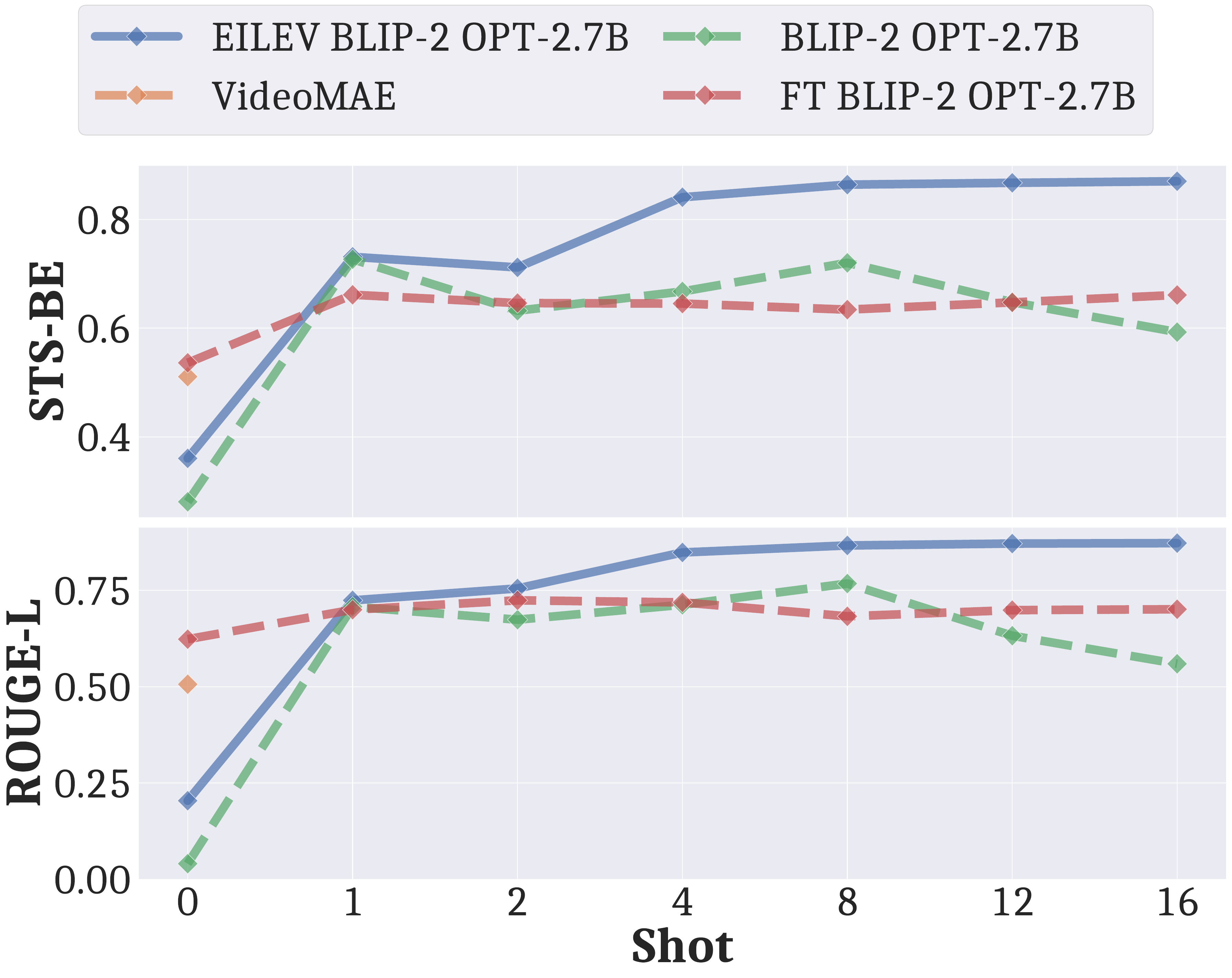}
    \end{subfigure}
    \begin{subfigure}{\linewidth}
        \centering
        \includegraphics[width=\linewidth]{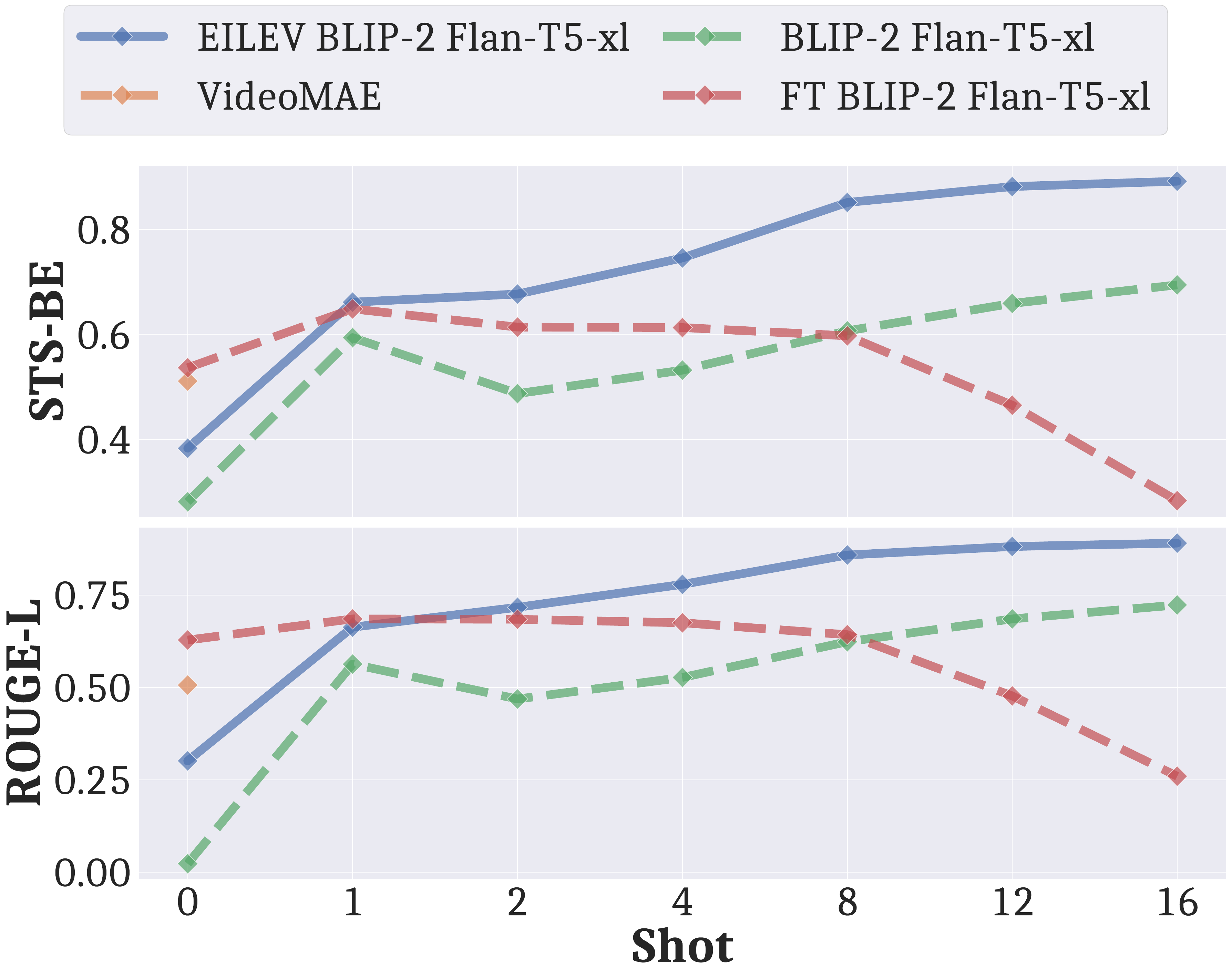}
    \end{subfigure}
    \caption{Performance of additional baselines on the EPIC-KITCHENS-100-based dataset}
    \label{fig:ek-eval}
\end{figure}

Figure \ref{fig:ek-eval} reports the performance of the three additional baselines on EPIC-KITCHENS-100. All the baseline models exhibit similar trends as on the Ego4D-based dataset: they demonstrate the best performance at 0-shot (``in-weights learning'') but fail to benefit from the in-context examples (``in-context learning''). 

\subsection{In-Context or In-Weights Learning}

We now aim to validate that the source of the generalization capabilities demonstrated by the \eilev-trained models in Section \ref{subsec:off-the-shelf-vlms} is indeed from in-context learning, not in-weights learning. This is to further reinforce our claim that \eilev-trained models can generalize to actions that they have not seen during training, i.e., actions of which they have no direct in-weights knowledge. To that end, we use the frequency of each verb/noun class in the common action training data as the proxy for the knowledge about the verb/noun class encoded into the weights of the model (in-weights learning), and the difference in model performance between 16-shot and 0-shot settings for a particular rare action as the proxy for in-context learning performance. If the model relies on in-weights learning for a particular novel, rare action, the difference in performance for that action between 16-shot and 0-shot settings would be correlated to the frequency of the corresponding verb/noun class in the training data. This outcome is not desired, as we want the model to rely on in-context learning for generating accurate narrations of novel, rare actions unseen during training.

\begin{figure*}
    \centering
    \includegraphics[width=\linewidth]{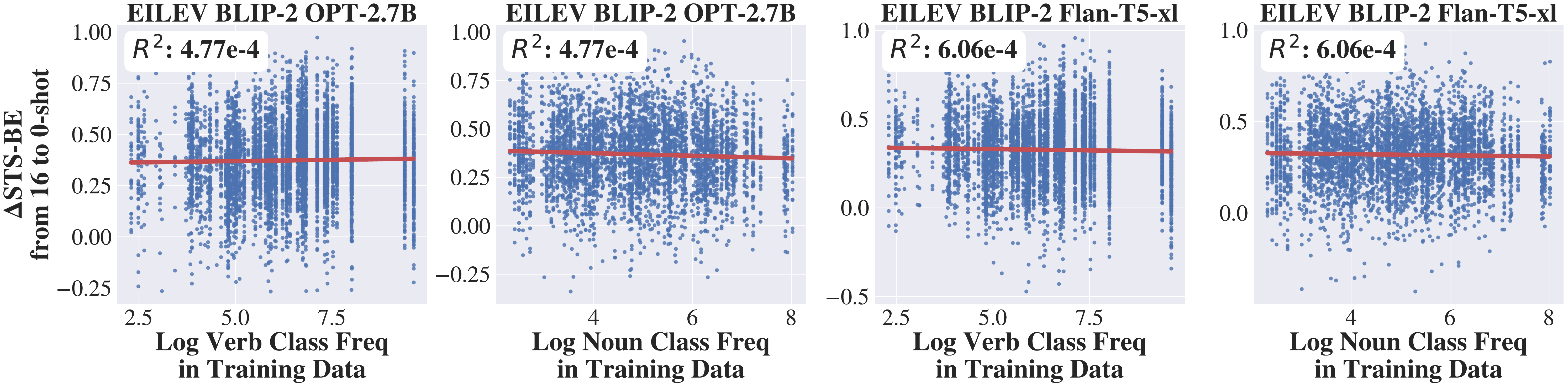}
    \caption{Scatter plots with trend lines and $R^2$ values between the log verb/noun class frequency in the training data with common actions and the difference in STS-BE ($\Delta$ STS-BE) for the corresponding rare action between 16-shot and 0-shot settings for the \eilev-trained models.}
    \label{fig:eval-freq-delta-sts-be}
\end{figure*}

Figure \ref{fig:eval-freq-delta-sts-be} shows the scatter plots between the log verb/noun class frequency in the training data and the difference in STS-BE for the corresponding rare action between 16-shot and 0-shot settings for the \eilev-trained models. For example, given a rare action (``put'', ``bench''), a point on the scatter plot may refer to the log frequency of ``put'' in the common action training data in the x-axis and the difference in the STS-BE performance of \eilev~BLIP-2 OPT-2.7B on (``put'', ``bench'') between 16-shot and 0-shot. As the scatter plots and their corresponding $R^2$ values show, there is a minimal linear correlation between the log verb/noun class frequency in the training data and the difference in STS-BE for the corresponding action from in-context learning. This suggests that the \eilev-trained models generate accurate narrations for novel, rare actions via in-context learning rather than in-weights learning, as the linear model does not significantly account for the variance in the observed data.

\subsection{Context Modeling and In-Context Learning}

In this evaluation, we seek to investigate if the \eilev-trained models perform correct context modeling by incorporating the relationships between video clips and narrations. To that end, we evaluate the \eilev-trained models and the off-the-shelf BLIP-2 baseline models from Section \ref{subsec:additional-baselines} on shuffled in-context examples where video clips no longer match the action narrations. We then compare their performance from shuffled in-context examples (the treatment group) to the one from un-shuffled in-context examples as the control group. If the performance remains unchanged, it implies that the model does not consider the relationships between in-context video clips and action narrations. On the other hand, if the performance decreases, it implies that the model does take the relationships between video clips and action narrations into account, and the mismatch adversely affects its performance. We do not report the results at 0 and 1-shot since shuffling of the in-context video clips would not have any impact at those settings.

\begin{figure}[ht]
    \centering
    \includegraphics[width=\linewidth]{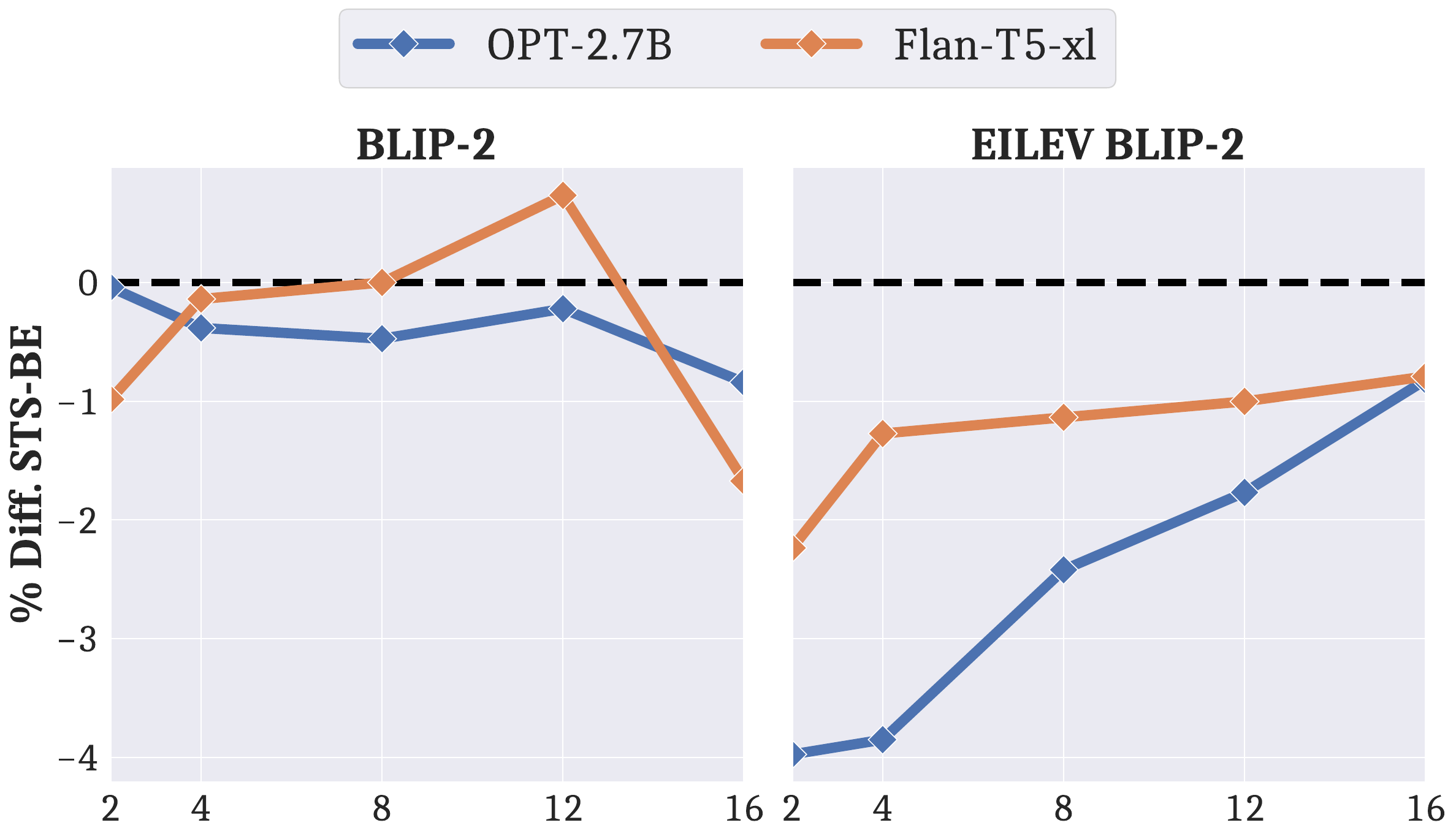}
    \vspace{-0.7cm}
    \caption{Percentage difference plots between the treatment group with shuffled in-context video clips and the control group. A negative value below the dotted zero line means the STS-BE performance of the treatment group is worse than the control group.}
    \label{fig:eval-shuffle-in-context}
\end{figure}

Figure \ref{fig:eval-shuffle-in-context} shows the percentage differences in STS-BE from 16-shot to 0-shot between the treatment group and the control group for the \eilev-trained models and the off-the-self BLIP-2 models. For the off-the-shelf BLIP-2 models, the percentage differences are small across all shots. This indicates that they rely mostly on the context as a whole rather than the semantic details from the relationships between video clips and action narrations when performing in-context learning. We hypothesize that our proposed architectural modifications (Section \ref{subsec:model} allow the off-the-shelf BLIP-2 models to tap into the text-only in-context learning capabilities of their frozen language models, which lack the ability to extract semantic details from the relationships between video clips and action narrations. This hypothesis is supported by their subpar in-context learning capabilities from Section \ref{subsec:additional-baselines}, which speaks to the importance of our modifications to the training data. On the other hand, there is a clear drop in performance for the \eilev-trained models in terms of the semantic-similarity-based metric STS-BE. This indicates that the \eilev-trained models extract detailed semantic information from the correspondence between in-context video clips and action narrations.
\section{Training Details}

In all of our experiments, each video clip is created by taking the four seconds before and after its action narration timestamp, and 8 frames are sampled uniformly from each video clip. The total training batch size is 128 and the optimizer is AdamW \cite{loshchilov2018decoupled} with the initial learning rate of $1\times10^{-5}$, weight decay of 0.05 and a linear scheduler. We train for 5 epochs on 8 NVIDIA A40 GPUs using distributed data parallel. We evaluate every 200 steps and select the model with the lowest loss. The training time is about a day and a half.
\section{Question Templates}
\label{sec:qa-template}

Table \ref{tab:qa-template} shows the question-answer pair templates we use in our experiments. They are based on the instruction templates proposed by \citet{dai2023instructblip}.

\begin{table*}
\caption{List of question-answer pair templates.}
\label{tab:qa-template}
\centering
\begin{tabular}{@{}l@{}}
\toprule
What is the camera wearer doing? \{narration\}                                                                                                        \\ \midrule
Question: What is the camera wearer doing? \{narration\}                                                                                              \\ \midrule
What is the camera wearer doing? An answer to the question is \{narration\}                                                                           \\ \midrule
Q: What is the camera wearer doing? A: \{narration\}                                                                                                  \\ \midrule
\begin{tabular}[c]{@{}l@{}}Given the video, answer the following question.\\What is the camera wearer doing? \{narration\}\end{tabular}               \\ \midrule
\begin{tabular}[c]{@{}l@{}}Based on the video, respond to this question:\\What is the camera wearer doing? Answer: \{narration\}\end{tabular}         \\ \midrule
\begin{tabular}[c]{@{}l@{}}Use the provided video to answer the question:\\What is the camera wearer doing? \{narration\}\end{tabular}                \\ \midrule
\begin{tabular}[c]{@{}l@{}}What is the answer to the following question?\\"What is the camera wearer doing?" \{narration\}\end{tabular}               \\ \midrule
\begin{tabular}[c]{@{}l@{}}The question "What is the camera wearer doing?" can be answered using the video.\\The answer is \{narration\}\end{tabular} \\ \bottomrule
\end{tabular}
\end{table*}

\end{document}